\documentclass[10pt,twocolumn,letterpaper]{article}

\usepackage{wacv}
\usepackage{times}
\usepackage{epsfig}
\usepackage{graphicx}
\usepackage{amsmath}
\usepackage{amssymb}
\usepackage{eso-pic}
\usepackage{booktabs} 
\usepackage{subfig} 

\wacvfinalcopy 

\def\wacvPaperID{****} 

\ifwacvfinal\fi
\begin{document}

\title{An Anti-fraud System for Car Insurance Claim Based on Visual Evidence}

\author{Pei Li \\
Univeristy of Notre Dame\\
{}
\and
BingYu Shen \\
University of Notre dame\\
{}
\and
Weishan Dong \\
IBM Research China
{}
}

\maketitle
\ifwacvfinal\thispagestyle{empty}\fi

\begin{abstract}
   Automatically scene understanding using machine learning algorithms has been widely applied to different industries to reduce the cost of manual labor. Nowadays, insurance companies launch express vehicle insurance claim and settlement by allowing customers uploading pictures taken by mobile devices. This kind of insurance claim is treated as small claim and can be processed either manually or automatically in a quick fashion. However, due to the increasing amount of claims every day, system or people are likely to be fooled by repeated claims for identical case leading to big lost to insurance companies.Thus, an anti-fraud checking before processing the claim is necessary. We create the first data set of car damage images collected from internet and local parking lots. In addition, we proposed an approach to generate robust deep features by locating the damages accurately and efficiently in the images. The state-of-the-art real-time object detector YOLO \cite{redmon2016you}is modified to train and discover damage region as an important part of the pipeline. Both local and global deep features are extracted using VGG model\cite{Simonyan14c}, which are fused later for more robust system performance. Experiments show our approach is effective in preventing fraud claims as well as meet the requirement to speed up the insurance claim prepossessing. 
\end{abstract}

\section{Introduction}

Fully or semi-automatically insurance claim processing could be very useful in the insurance industry when handling small but more frequent insurance claims that under a certain amount of rate. It increases the speed of claim investigation and loss assessment. Some of the companies employ 3G mobile techniques to accelerate the investigation process by freeing the customers from submitting complicated claim proves in order to shorten the whole claim process up to eight hours. However, fraud claims for which the payment has been claimed more than one time could potentially cause loss to insurance companies. Manually validating on large scale of claims cannot meet the speed requirement for express claim process anymore and reporting a duplicated claim candidates automatically before the settlement is needed. However, this solution remains a challenging task due to a number of factors: In pictures or video frames captured by mobile devices, the viewpoint and illumination conditions are not taken in a controlled environment; Damage cannot be localized accurate and fast enough to generate robust local features for matching; No standard data set is available for analyzing this problem. We made two major contributions to address the problem: 
\begin{itemize}
  \item we contribute the first car damage dataset which contains samples collected from both internet and local public parking lot. In addition, manually annotation is provided.
  \item We proposed an anti-fraud system prevent repeated claims happens before the settlement is issued. 
\end{itemize}

The whole system consists of two parts: a real-time damage detector to provide accurate damage locations in the picture and deep feature extractor to generate combined global and local deep features for anti-fraud matching in claim history database. We fine-tuned our data on the real-time object detector YOLO which reported to have 457FPS on VOC2007 with a mAP of 63.4 with its standard model. To handle the affine transformation of the damages in images due to viewpoint changes, the possible various lighting condition and the shift caused by the detection result, we tried to employ local respond normalization layer and dropout as well as data augmentation in the training process to make the model more generalized. Our system is evaluated with both data set collected on internet from three of the biggest searching engines and local public parking lot.
\section{Related Works}
\subsection*{General Object Detection and Damage Detection}
Accurate localization of the damage on a car plays an important role in our system.General object detection is a important topic in computer vision. Classical detection pipeline will first conduct a sliding-window fashion candidate region selection. Then either hand-craft feature\cite{papageorgiou1998general}\cite{lowe1999object}\cite{dalal2005histograms} or deep features will be extracted from each of the candidate regions for binary classification that tells if it's a detection or not. Later, Deformable parts models(DPM) is introduced to run detection with detectors trained on specific regions of the image in an exhaustive manner across all locations and scales.The exhaustive search through all possible locations and scales poses a computational challenge and big time delay on each detection. Recently, some detection frameworks replaced the sliding windows with region proposals generated either from selective search or from feature map bonding box prediction, such as \cite{girshick2015fast} \cite{ren2015faster} \cite{liu2016ssd} \cite{girshick2016region}. Another way of doing object detection is to treat the problem as a bounding box regression problem in general \cite{erhan2014scalable} \cite{redmon2016you}, but among all of them, YOLO is the only one that can meet real-time requirement.
Damage detection is a core problem during quality control. Several damage detection approaches have been proposed applied to car body damage detection. Srimal.al\cite{jayawardena2013image} propose to use 3D CAD models to handle automatic vehicle damage detection via photograph. Gontscharov.al \cite{gontscharov2014algorithm} tries to solve vehicle body damage multi sensor-data fusion. Keyence Vision\cite{WinNT} proposed an industrial solution for car damage by hail by applying a high-resolution Multi-camera vision system.  Cha.al.\cite{cha2017deep}adopt image based deep learning to detect crack damages in concrete while \cite{cha2017output} adopted a phase-based optical flow and unscented Kalman filters.summary of the up-to-date crack detection using image processing\cite{mohan2017crack} which is similar to car damage detection.

\section{Dataset}
\begin{figure}[t]
\begin{center}

   \includegraphics[width=0.8\linewidth]{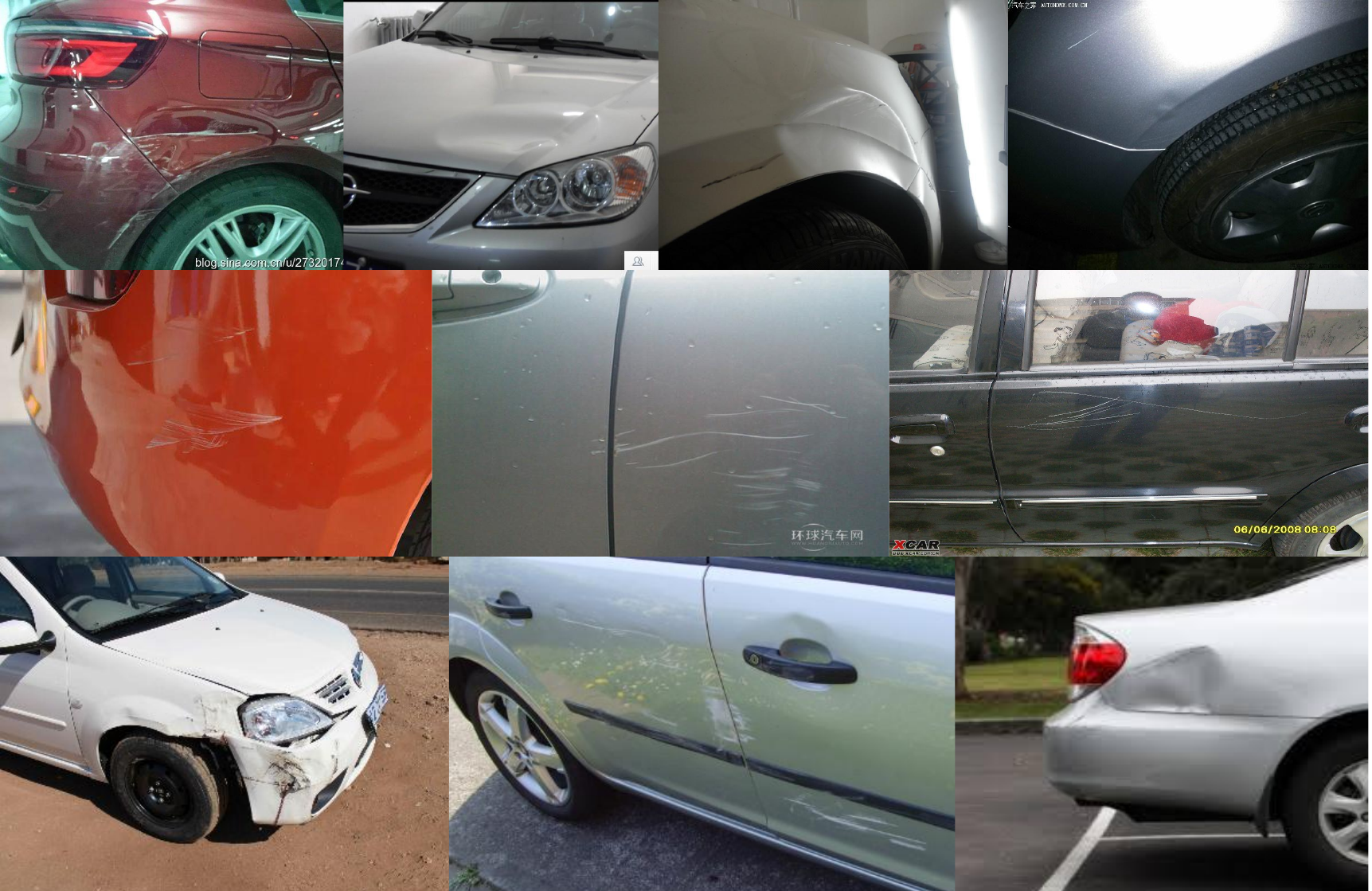}
\end{center}
   \caption{Example images from images collected from the internet}
\label{fig:damage_sample}

\end{figure}

\begin{figure}[t]
\begin{center}

   \includegraphics[width=0.8\linewidth]{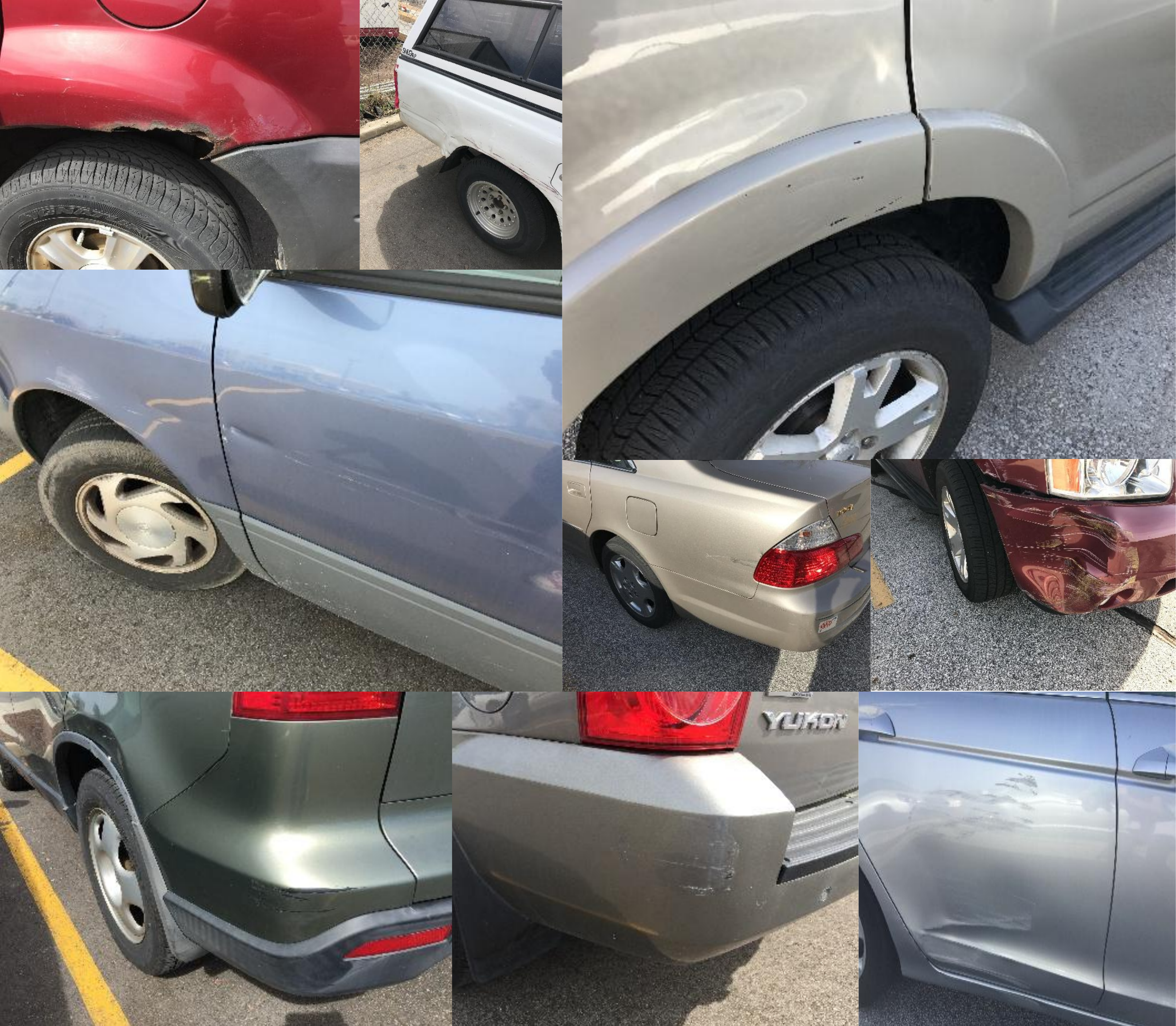}
\end{center}
   \caption{Example images from images collected from a local public parking lot}
\label{fig:probe_gallery_exp}

\end{figure}

We introduced a car damage dataset collected from one of three biggest searching engines: Google Bing and Baidu using keywords like 'scratch','dent'. This contains 1790 images with manual annotation. In addition, we collect vehicle images with small damage in public parking lot without including the identification information of the vehicles.These data set has 92 cars each with four or five images and are also manually annotated with bounding boxes tightly bounds the damage regions. In order to mimic customer behaviors, we use mobile devices to capture the most convincing evidence for a successful insurance claim. In each case, we capture images of a whole car body and local scratch images from the same vehicle. Each of the cases contains two images captured the same damage regions which represent the first claim and the other represents the repeated claim that treated as a fraud claim. Example images are shown in Figure \ref{fig:damage_sample} and Figure \ref{fig:probe_gallery_exp} with random samples from the dataset. The whole dataset is used for two purposes: one for damage detection training and evaluation, another for studying the whole system which contains both damage localization and fraud claim detection.

\section{The anti-fraud system for car insurance claim}
\begin{figure*}[ht]
\label{fig:process}
\begin{center}

   \includegraphics[width=12cm]{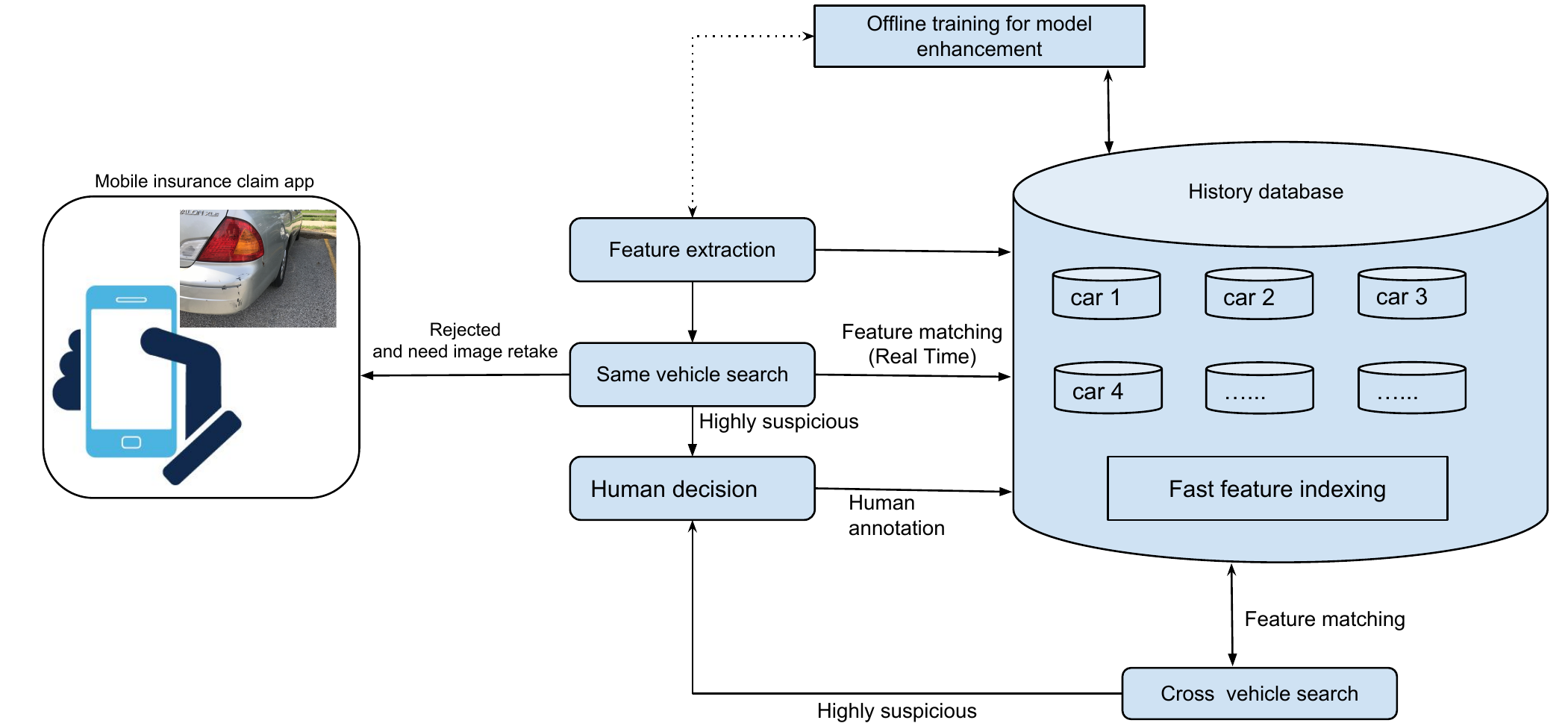}
\end{center}
   \caption{Anti-fraud system overview}
\label{fig:system overview}

\end{figure*}

The anti-fraud or fraud detection system is coupled with the claim history database. Each user has their own record in the database after they have a claim for the first time. The fraud claim could happen in two ways: the same-vehicle reclaim or cross-vehicle reclaim. Same-vehicle reclaim happens when the user tries to make a quick insurance claim by uploading a similar image which is collected during the same case. Another fraud claim would happen when the user uses a similar image taken from another car as his or her own vehicle claim. Both formats could lead to reclaim for the same cases actually has been issuing the settlement. Traditional insurance claim is handled manually by representatives which is necessary when the claim value is above a certain threshold. However, when the claim number increasing each day and some of the claims are only about small damages such as scratches dent etc. Manually handle all these claims decrease the efficiency of insurance companies' service and increase the cost of staffing. In this case, a system that can help to detect the fraud claims is needed when we want to make the system more automatically. As shown in Figure \ref{fig:process}, the user is required to upload several images following the requirement in order to open a new claim. The anti-fraud system will then using algorithms to extract features to search in the enrolled history database. If the system reports a high score of suspicious of a certain claim, humans can take into the loop and manually retrieve the history to do the loss assessment. To make the system more accurate and fast, we need a robust feature which based on accurate locating the damages in the image. We adopt two state-of-the-art real-time detectors for damage detection and propose an approach to form robust features for the anti-fraud searching. We will introduce each part in the following sections.

\subsection*{Damage detection}
In most of the cases, users are required to upload few images of the close view of the damages on the vehicle and another picture with a farther view that contains the whole body of the car as well as the plate which shows the model and identity of the car. The first step to obtaining a robust feature descriptor for fraud detection is the damage detection. This task is challenging since it is not a traditional object detection problem and the damages would have different forms. We consider the damage mainly three types due to analyzing our dataset: scratch, dent, and crack.

Localization of a damage in an image uploaded by users is a critical part of our system as the system would obtain more stable and discriminative features when accurate detection happens. In addition, to meet our real-time requirement without sacrifice detection accuracy, we employed YOLO, which is a state-of-the-art lightweight real-time detector to be part of the system. We proposed our approach to train and evaluate on the new dataset we collect.
\subsubsection{YOLO Detector}
YOLO is a fast, accurate object detection framework.Its base network runs at 45 frames per second with no batch processing on a Titan X GPU and a fast version runs at more than 150 fps which allows us to process streaming video in real-time with less than 25 milliseconds of latency. It is extremely useful to assist needs in auto-drive, real-time responsive services.
YOLO frame object detection as a single regression problem.Image pixels are mapped to bounding box coordinates and class probabilities straightly. As damages like scratch or dent might have a different shape, YOLO overcome this by learning features through regression of four coordinates.YOLO, with its standard model, achieve 45FPS with mAP of 63.4 on VOC2007. All these performances meet the requirement of the efficient needs of an express anti-fraud system for insurance claims.

\subsection*{Damage Re-Identification}
After damage detection generated, robust features need to be extracted to overcome lighting conditions and different angle of views. Also, both global and local features are being used to finally determine whether a claim has ever appeared in the history database as each claim requires at least one image of the image containing the close view of the damage as well as another image shows the whole view of the body of the corrosponding vehicle . To extract local features, we employ the pre-trained VGG16 \cite{Simonyan14c}object recognition model as a feature extractor. Global feature consists of global deep features and color histogram. We fuse these feature to obtain a more discriminative feature and conduct a 1 to N match between the probe images and the images in the post-claim history database.

\section{Experiments and result}

\subsection{Damage detection}
\begin{figure}[ht]
\begin{center}

   \includegraphics[width=0.8\linewidth]{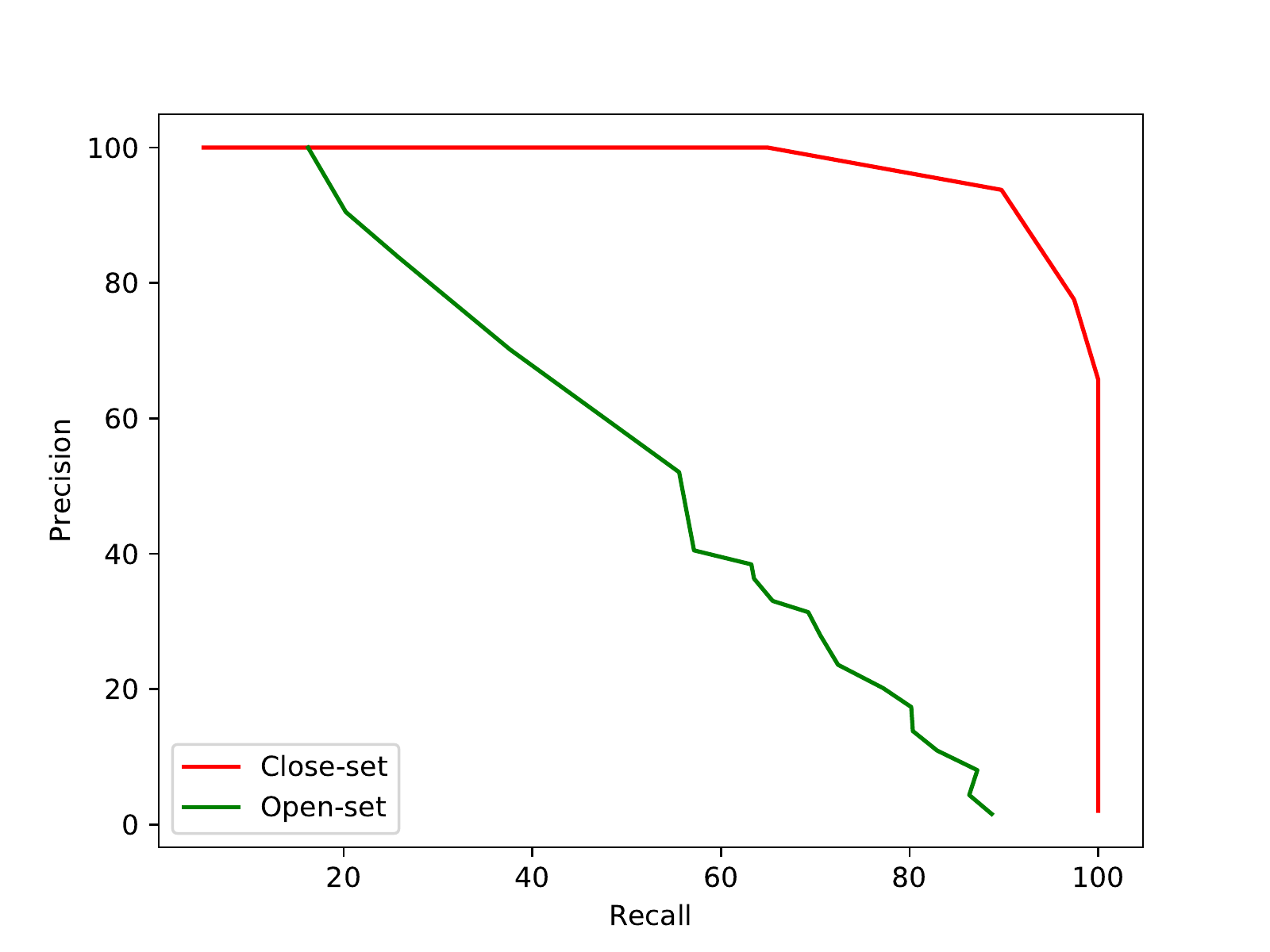}
\end{center}
   \caption{Detection result on real data collected in the wild}
\label{fig:closeopen}

\end{figure}

\begin{figure}[ht]
\begin{center}

   \includegraphics[width=0.8\linewidth]{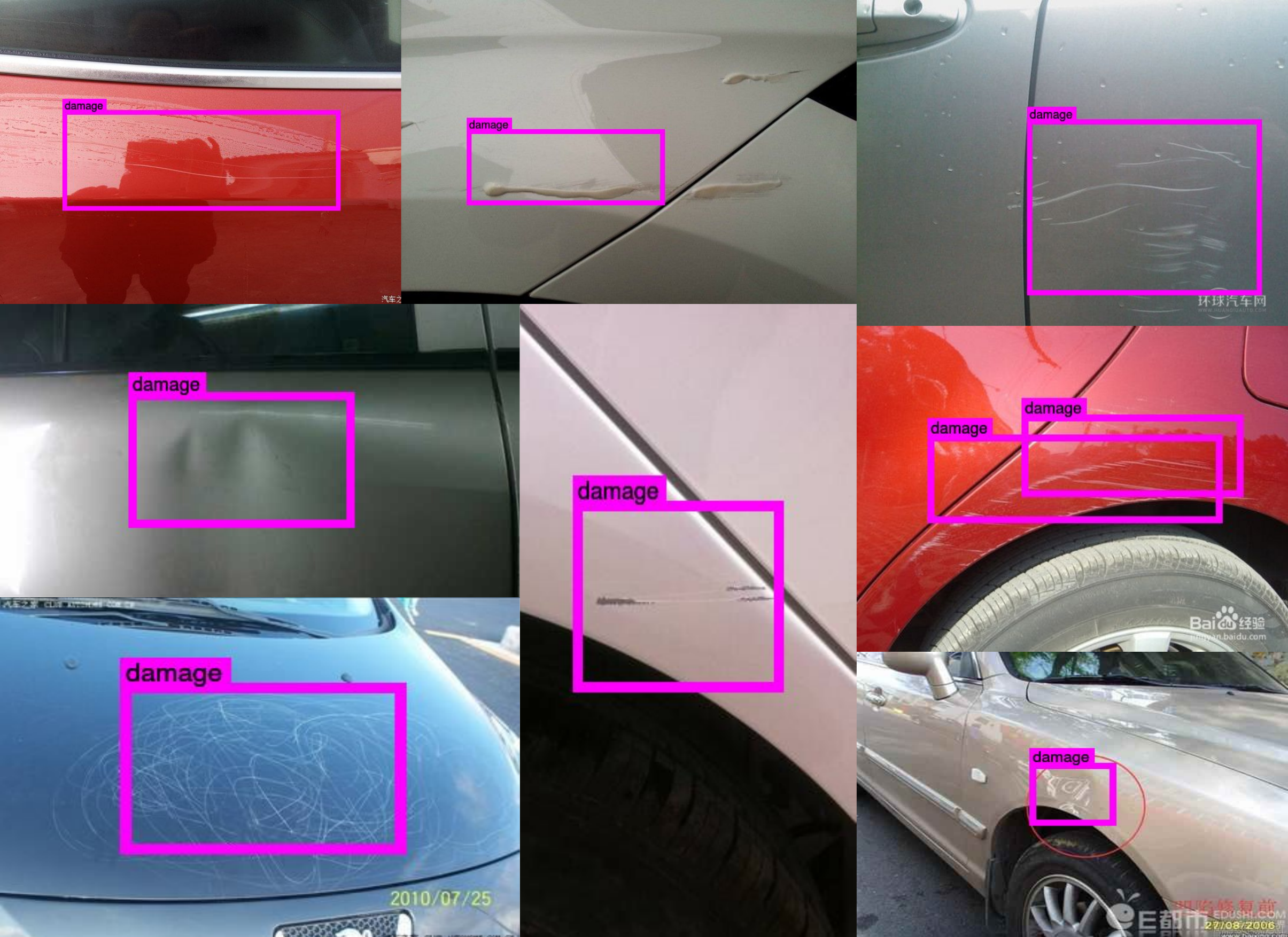}
\end{center}
   \caption{Detection examples on data collected online}
\label{fig:dmage_detection_wild}

\end{figure}
To cope with the size of the data set, we select different pre-trained models in YOLO and conduct a transfer learning by fine-tuning with our data. This contains two sets of experiments: Training and evaluation on the web-collected dataset and evaluation on the real car damage data set from a real parking lot.  For the first set of experiment, we use around 1300 images for training which come from Google and Bing and 400 images for testing which are from Baidu. We evaluated on different network structures starting with the YOLO-tiny model which is the simplest and fastest network structure among the models reported in \cite{redmon2016you}. We later tried to add dropout layers and local respond normalization layers aiming to handle the different lighting condition and reduce overfitting on our small scale data set.
From the experiment result shown in Table\ref{pr_on_baidu}, we can see YOLO with LRN layer perform the best in general achieved 81.7 percent of recall and 37.96 percent of precision at threshold 0.1. The YOLO with LRN layer and dropout layer achieve better precision at 47.04 percent while lower recall. The VGG-16 has a lower performance than the YOLO original model and the Tiny YOLO model achieves the lowest performance due to the size of the net architecture.

Later, we conduct an experiment over real car damage data set which is collected in the local parking lot. We designed two protocols for the training and testing. Consider the real system is updated periodically enhancing the model, all the images recorded as history in the system could be used for training the model. As shown in \ref{fig:process}. In this case, we consider this evaluation protocol subject-overlapped detection. Another evaluation protocol follows the previous experiment which the training and testing set are separated by car ID. The precision-recall curve shows in \ref{fig:closeopen} indicates that the gap between subject-overlapped and subject-disjoint detection result. With more relevant images used for training the detector, the more accurate a similar damage could be detected and as the size of the  history database increase, better performance of the detector is expected .

\begin{table*}[htbp]
  \centering
  \caption{Precision Recall @thresholds}
	\label{pr_on_baidu}
    \begin{tabular}{|l|r|r|r|r|r|}
    \toprule
          & \multicolumn{1}{l|}{YOLO\_original} & \multicolumn{1}{l|}{YOLO\_vgg\_16} & \multicolumn{1}{l|}{YOLO\_with\_LRN} & \multicolumn{1}{l|}{YOLO\_with\_LRN+dropout} & \multicolumn{1}{l|}{YOLO\_tiny} \\
    \midrule
    precision & 12.66 & 11.1  & 14.62 & 14.27 & 9.41 \\
    \midrule
    recall & 85.6  & 82.01 & 89.72 & 78.66 & 72.49 \\
    \midrule
    precision & 25.58 & 32.75 & 37.96 & 47.04 & 17.48 \\
    \midrule
    recall & 76.61 & 57.58 & 81.75 & 63.54 & 66.84 \\
    \midrule
    precision & 40.89 & 49.07 & 38.63 & 86.25 & 26.26 \\
    \midrule
    recall & 66.32 & 27.25 & 72.49 & 17.74 & 60.15 \\
    \bottomrule
    \end{tabular}%
  \label{tab:addlabel}%
\end{table*}%


\begin{figure*}[ht]
\centering

   \subfloat{{\includegraphics[height=5.5cm,width=7cm]{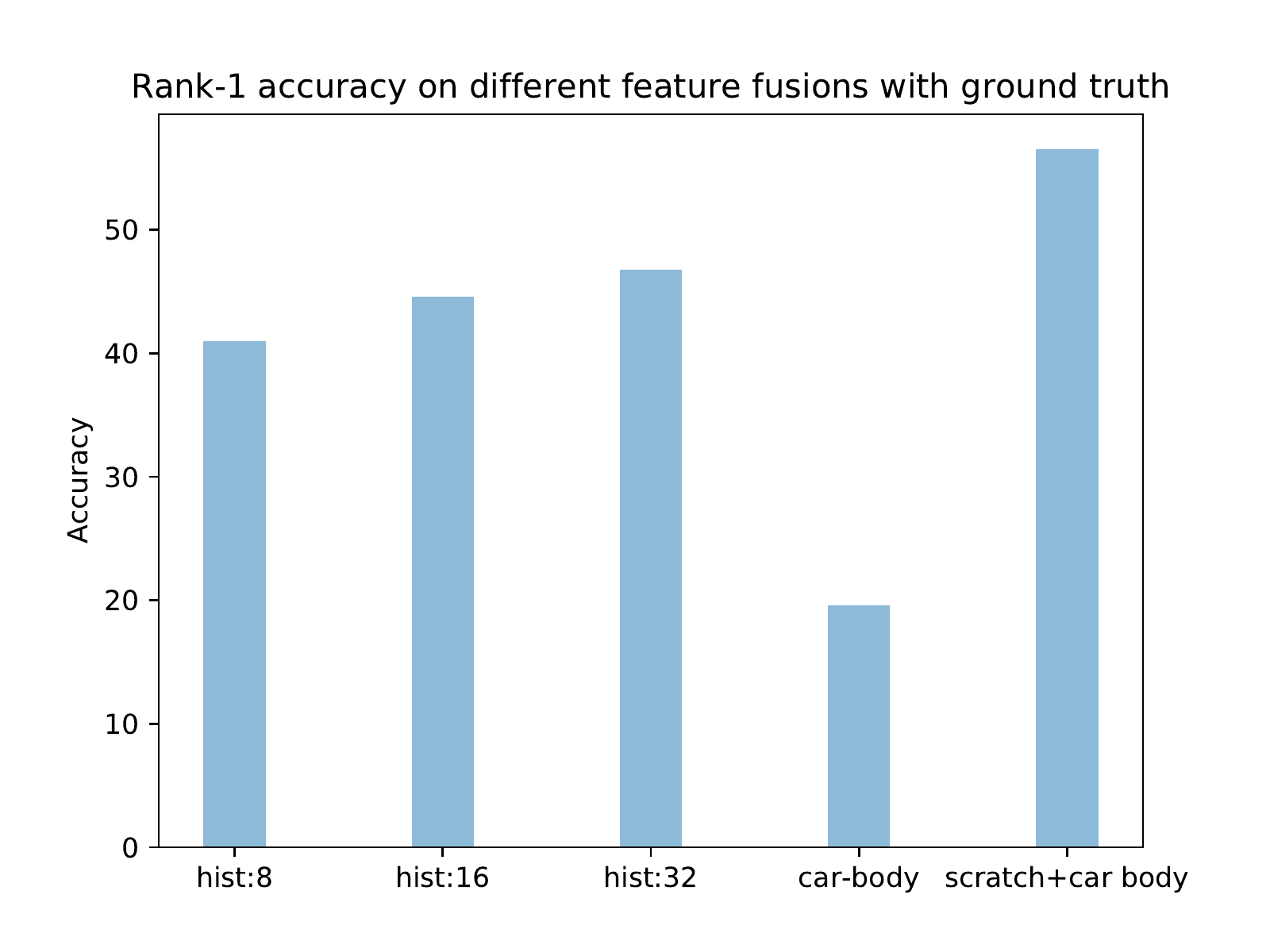} }} 
 \subfloat{{\includegraphics[height=5.5cm,width=7cm]{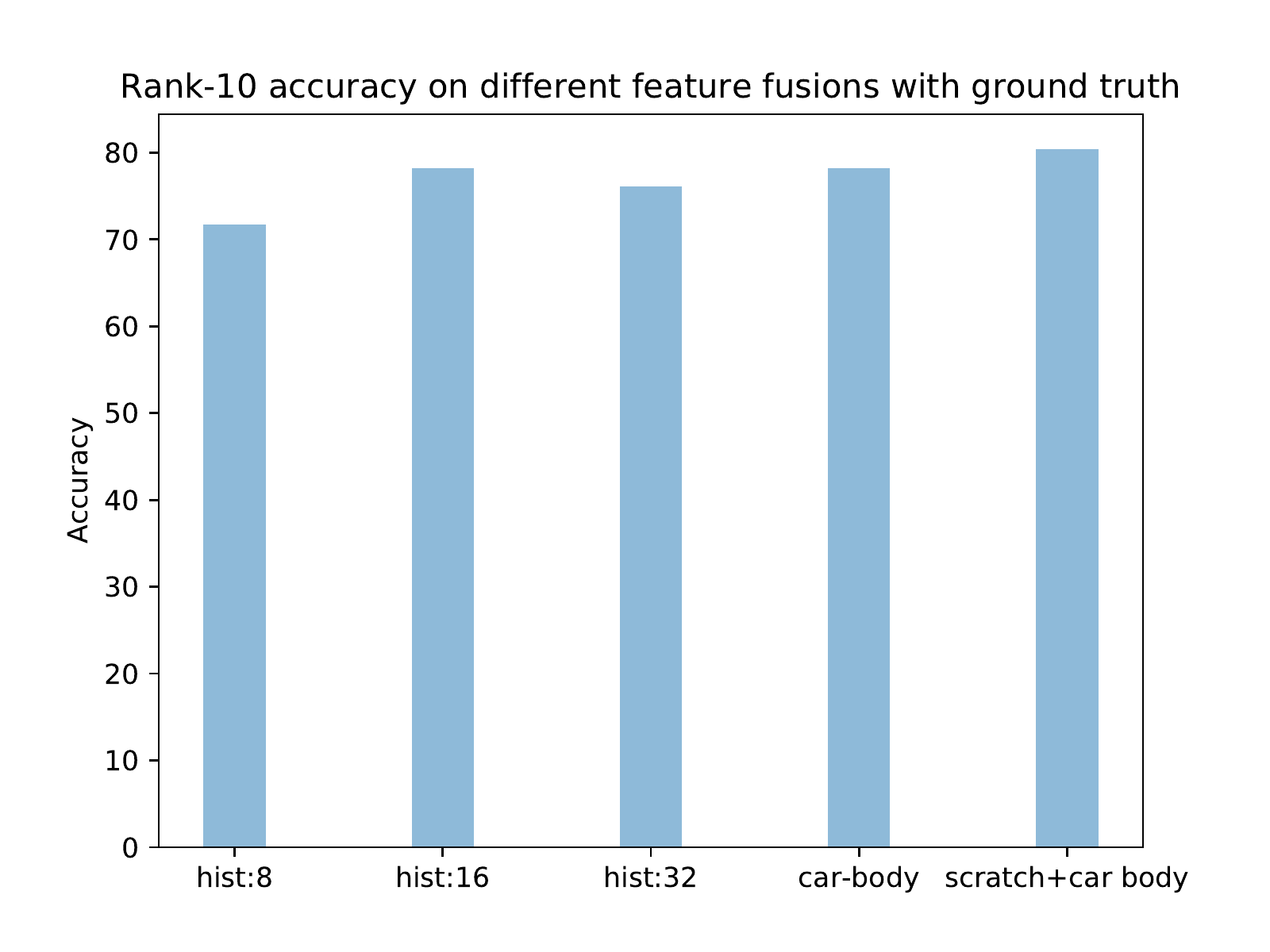}  }}
    
    \subfloat{{\includegraphics[height=5.5cm,width=7cm]{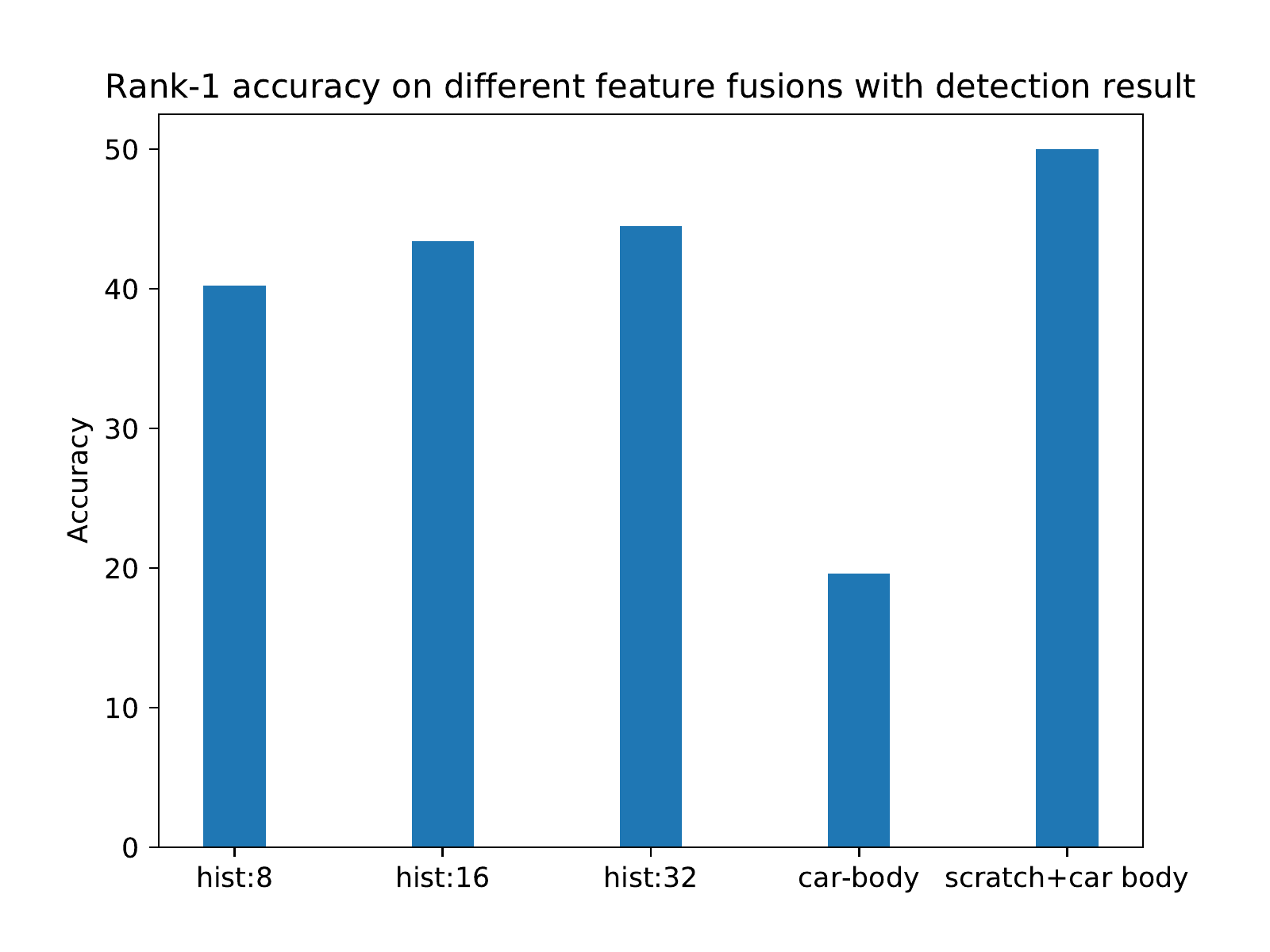} }}
    \subfloat{{\includegraphics[height=5.5cm,width=7cm]{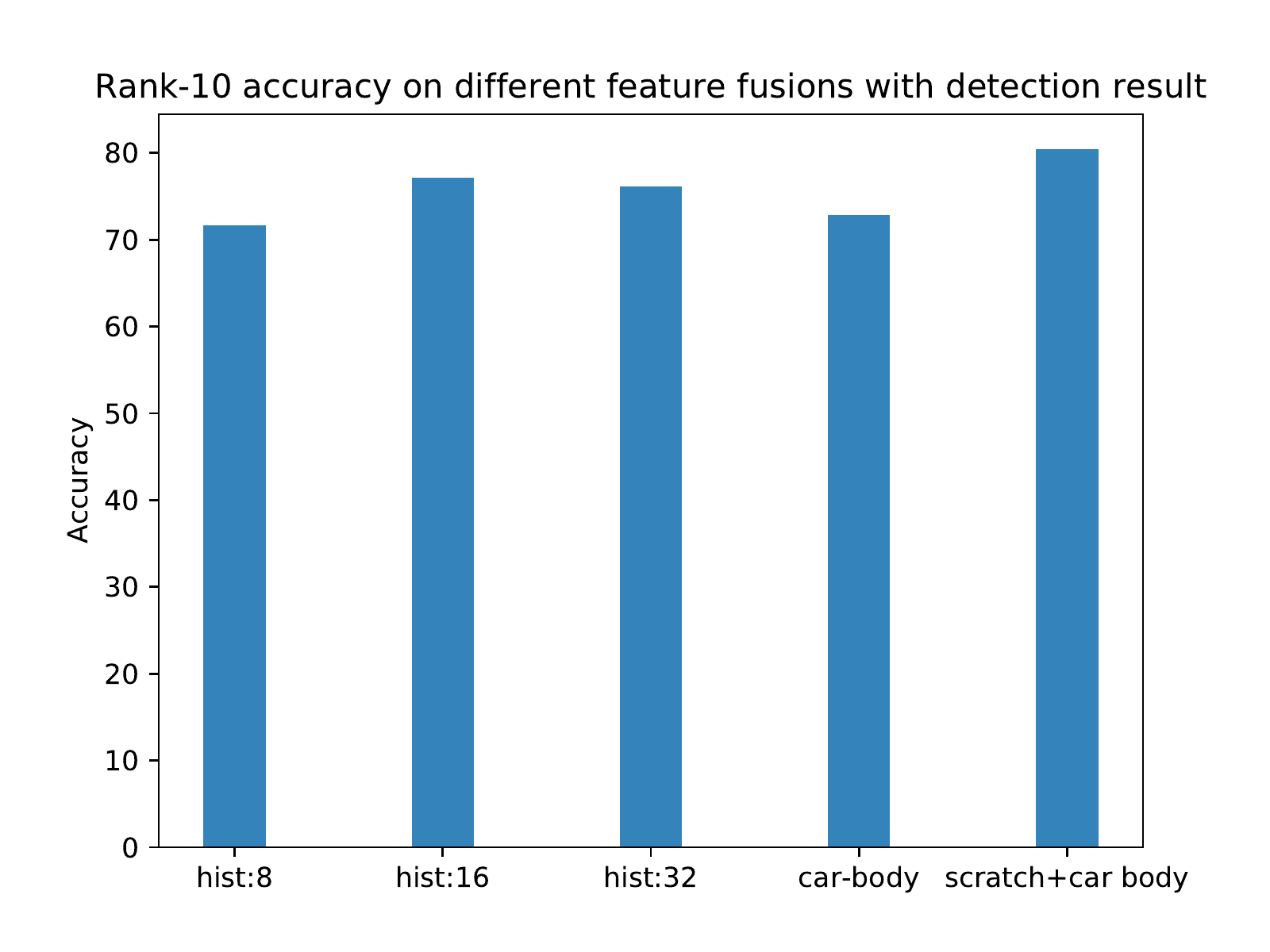} }}
   
   \caption{Fraud detection result at rank-1 rate and rank-10 rate with ground truth and detection result with different feature fusions}
\label{fig:rank}
\end{figure*}
\subsection{Fraud claim detection}
In this section, we design experiment aiming to mimic the real fraud cases. Under this scenario, the history database contains at least one images that has been claimed and issued settlement. We consider the most general case which is the cross vehicle search. It means that we have to search across the whole database not only the history belong to a certain user. We random chose one damage image per vehicle as probe and the rest as gallery. To get a more robust feature for fraud detection, we exploit feature fusion combining local feature, global feature as well as histogram feature to obtain a final feature descriptor. Local feature is extracted from the damage region which is detected by the damage detector. Global feature is extracted from the corresponding full vehicle image which could potentially help to make the feature more discriminative since the damages could be very similar from different claims. We also incorporate histogram feature as a global feature expecting to enhance the final descriptor. All the features from the scratch and car body are extracted from the last FC layer of the pre-trained VGG-16 model and concatenate to form the final descriptors. Cosine distance is used to compare each probe to the gallery images.Rank-1 and Rank-10 rate are reported as shown in \ref{fig:rank}. We also study how different dimension of the histogram feature affect the performance. We firstly used the manually annotated damage region as input which we assume that the damage detection is perfect. After this experiment, we From the experiment result shown in \ref{fig:rank} we gain a 41.0 percent rank-1 rate with 8 bins in each channel of the histogram feature. When we increase the bin to 16 and 32, the performance increase to 44.6 percent and 46.73 percent. For the rank-10 performance， increasing the dimension of the color histogram feature does not always boost up the performance. When adopting the 32 bin color histogram feature, a slightly performance drop happens.
However, if only employ the global feature obtained from the whole vehicle body, we only gain 19.6 percent rank-1 rate, while incorporating the feature from the scratch enhance the rank-1 rate to 56.52 percent. This indicates that the histogram feature might have negative impact on the performance when the color of the candidate vehicles are very similar.

We also explore the matching performance by substitute the detection result from the ground truth with the ones yield from the YOLO detector we trained. We discover only a tiny performance difference when directly employ the detector which indicates that the detector if trained well could yield a robust result for the damage feature extraction.

\section{Conclusion and discussion}

In this paper, we introduce a useful data set of car damage which is the first data set for car damage detection and matching. We proposed a practical system pipeline to detect the fraud claim before issuing the settlement in order to reduce loss for the insurance company. We research on different deep architectures for better damage detection result and analyze different feature fusions and its impact on the final matching performance.Extensive experiments with different experiment settings under real scenario demonstrate the effectiveness of the proposed approach.

{\small
\bibliographystyle{ieee}
\bibliography{egpaper_final}
}

\end{document}